\documentclass[sigconf]{acmart}
\usepackage{amsmath}
\usepackage{graphicx}
\usepackage{algorithm}
\usepackage{algpseudocode}
\usepackage{enumitem}
\usepackage{multirow}
\setlist{leftmargin=5mm}
\usepackage{booktabs}
\usepackage{caption}
\AtBeginDocument{%
  \providecommand\BibTeX{{%
    \normalfont B\kern-0.5em{\scshape i\kern-0.25em b}\kern-0.8em\TeX}}}

\setcopyright{acmlicensed}
\copyrightyear{2024}
\acmYear{2024}
\acmDOI{XXXXXXX.XXXXXXX}

\acmConference[Conference acronym 'XX]{Make sure to enter the correct
  conference title from your rights confirmation emai}{June 03--05,
  2024}{Woodstock, NY}
\acmISBN{978-1-4503-XXXX-X/18/06}

\begin{document}

\title{DUE: Dynamic Uncertainty-Aware Explanation Supervision via 3D Imputation}



\author{Qilong Zhao}
\affiliation{%
  \institution{Emory University}
  \country{USA}}
\email{qzhao31@emory.edu}

\author{Yifei Zhang}
\affiliation{%
  \institution{Emory University}
  \country{USA}}
\email{yifei.zhang2@emory.edu}

\author{Mengdan Zhu}
\affiliation{%
  \institution{Emory University}
  \country{USA}}
\email{mengdan.zhu@emory.edu}

\author{Siyi Gu}
\affiliation{%
  \institution{Stanford University}
  \country{USA}}
\email{sgu33@stanford.edu}

\author{Yuyang Gao}
\affiliation{%
  \institution{The Home Depot}
  \country{USA}}
\email{yuyang_gao@homedepot.com}

\author{Xiaofeng Yang}
\affiliation{%
  \institution{Emory University}
  \country{USA}}
\email{xyang43@emory.edu}

\author{Liang Zhao}
\authornote{Corresponding author}
\affiliation{%
  \institution{Emory University}
  \country{USA}}
\email{liang.zhao@emory}

\renewcommand{\shortauthors}{Zhao, et al.}

\begin{abstract}

Explanation supervision aims to enhance deep learning models by integrating additional signals to guide the generation of model explanations, showcasing notable improvements in both the predictability and explainability of the model. However, the application of explanation supervision to higher-dimensional data, such as 3D medical images, remains an under-explored domain. Challenges associated with supervising visual explanations in the presence of an additional dimension include: 1) spatial correlation changed, 2) lack of direct 3D annotations, and 3) uncertainty varies across different parts of the explanation. To address these challenges, we propose a Dynamic Uncertainty-aware Explanation supervision (DUE\footnote{Code available at: \href{https://github.com/AlexQilong/DUE}{https://github.com/AlexQilong/DUE}.}) framework for 3D explanation supervision that ensures uncertainty-aware explanation guidance when dealing with sparsely annotated 3D data with diffusion-based 3D interpolation. Our proposed framework is validated through comprehensive experiments on diverse real-world medical imaging datasets. The results demonstrate the effectiveness of our framework in enhancing the predictability and explainability of deep learning models in the context of medical imaging diagnosis applications. 

\end{abstract}

\ccsdesc[500]{Computing methodologies~Supervised learning}
\ccsdesc[500]{Computing methodologies~Computer vision}

\keywords{Explanation Supervision, 3D Data, Uncertainty Quantification, Visual Explanation}

\maketitle
\pagestyle{plain}

\section{Introduction}
While deep learning models demonstrate exceptional performance in computer vision, their ``black box'' feature raises concerns about their application in high-risk areas. In constructing transparent and trustworthy models, Explainable AI (XAI) has become a critical focus, especially in medical imaging. Existing works have introduced various XAI techniques, such as saliency maps~\cite{selvaraju2017grad,sundararajan2017axiomatic, miglani2023using,kindermans2016investigating}, which elucidate the features responsible for model predictions. However, there has been limited attention devoted to the quality of model explanations, including their fidelity to predictions and strategies for enhancing explainability when ground truth explanations are absent or inaccurate. Beyond traditional XAI techniques, an emerging research direction known as \textit{explanation supervision} aims to incorporate additional supervision signals during the learning process of a model,  in improving both the generalizability and explainability of deep learning models. 


Current methods for explanation supervision have been extensively examined across tabular data, natural language, and two-dimensional (2D) image data. For tabular and natural language data, existing studies~\cite{arrieta2020explainable, adadi2018peeking, hajialigol2023xai} have leveraged techniques such as attribution and feature regularization as means of supervision to enhance models. For 2D image data, recent studies~\cite{shen2021human, gao2022res, gu2023essa} focus on jointly optimizing explanation loss and prediction loss, by comparing human explanation annotations with model-generated saliency maps from post-hoc \cite{montavon2019layer, simonyan2013deep} or intrinsic explainers~\cite{freitas2014comprehensible,ribeiro2016model} along with comparing the predicted label and ground-truth label. However, the research into the direct application of explanation supervision techniques to 3D data remains an under-explored domain. This gap is notable given the abundance of 3D data in real-world applications, particularly in the field of medical imaging, where images such as computed tomography (CT) scans and magnetic resonance imaging (MRI) intrinsically present data in a 3D format.

This lack of exploration can be attributed to various fundamental challenges associated with supervising visual explanations when an additional dimension is involved:
\textbf{1) Spatial correlations change from 2D to 3D.} The shift from 2D to 3D image data induces a significant change in spatial correlation, as the additional dimension introduces depth. Unlike 2D image data, where spatial information is confined to width and height dimensions, and each annotation slice is treated as independently distributed, the intricacy of 3D image data necessitates modifications to both the model architecture and the explanation supervision paradigm. The model must capture spatial features and correlations in the third dimension, potentially causing misalignment between data patterns and the learning capabilities of the paradigm. 
\textbf{2) Gaps between 2D explanation annotations and 3D images.} Humans usually cannot directly delineate a precise curvature surface for 3D complex objects in 3D space. Instead, it is intuitive to label annotations on a few 2D slices (usually with a limited number of them to constrain the labor cost). Hence, such 2D slices cannot fully represent 3D explanation annotations, and leads to a gap when being used for explanation supervision on 3D images. This gap impedes effective explanation-guided learning in fields like medical imaging, where training samples are often limited.
\textbf{3) The quality of annotations varies in 3D space.} 
The \emph{curse of dimensionality} tells us that 3D space is ``much larger'' than 2D space, making it almost impossible to maintain the explanation annotations to have the same quality at any point in 3D space. Therefore, it is very important to identify the quality of explanation annotations to customize the strength of supervision accordingly. However, automatic estimation of the quality of explanation annotation is extremely difficult.


To address above challenges, we propose a \textbf{\underline{D}}ynamic \textbf{\underline{U}}ncertainty-aware \textbf{\underline{E}}xplanation supervision (\textbf{DUE}) framework for 3D explanation supervision that ensures uncertainty-aware explanation guidance when handling sparsely annotated 3D data. The uncertainty-aware guidance is achieved by integrating a 3D explanation loss term, a diffusion-based distance-sensitive interpolation method, and a post hoc weighting module. This module dynamically fine-tunes the weights assigned to the smallest individual units in the interpolated annotation slices based on their respective levels of uncertainty. 

Specifically, our main contributions are summarized as follows:

\begin{enumerate}
\item \textbf{Proposing a DUE framework for explanation supervision in 3D.} We propose a novel framework that extends the application of explanation supervision to the 3D domain, thereby improving the predictability and explainability of 3D models.

\item \textbf{Introducing a module for uncertainty-aware guidance.} Our approach introduces an uncertainty quantification module that dynamically estimates uncertainty levels. These estimations are utilized to weight the interpolated annotation slices, providing uncertainty-aware explanation guidance.

\item \textbf{Proposing an objective for incomplete and uncertain 3D annotations.} We propose an explanation loss term to handle the challenges introduced by incomplete 3D annotations and noisy interpolation. The computation of this loss term involves interpolated annotation slices and their weights.

\item \textbf{Conducting comprehensive experiments to evaluate our proposed approach.} We conduct comprehensive experiments on various real-world datasets and employ diverse evaluation metrics, demonstrating the effectiveness of our approach. Additionally, we present a thorough analysis of the generated explanations, showing their consistency and informativeness.
\end{enumerate}

The rest of the paper is organized as follows: Section~\ref{sec:background} reviews the background and related work, and Section~\ref{sec:problem_formulation} presents the problem formulation. Section~\ref{sec:method} describes the proposed DUE framework. The experiments on 2 real-world tasks are provided in Section~\ref{sec:experiment}, and the paper concludes with a summary of the research in Section~\ref{sec:conclusion}

\section{Related Work} \label{sec:background}

\subsection{Images Interpolation}
Image interpolation is a fundamental task of image processing to enlarge an image’s size or resolution. Traditional techniques like linear and cubic interpolation~\cite{maeland1988comparison} have been foundational but often insufficient for capturing the complex details necessary for accurate medical diagnoses. Methods such as Neighbor Mean Interpolation~\cite{huang2015reversible}, Interpolation by Neighboring Pixels~\cite{zhang2011reversible}, and New Interpolation Expansion~\cite{hong2011reversible} take one step forward in improving detail preservation and accuracy by leveraging local pixel relationships more effectively. While works like Marching Cubes~\cite{lorensen1998marching} and Volume Rendering~\cite{drebin1988volume} focus on interpolating 3D data. Recent years have witnessed a shift towards machine learning approaches, particularly with Convolutional Neural Networks (CNNs) making notable advancements in preserving anatomical structures more effectively~\cite{mao2019interpolated}. Oring et al.~\cite{oring2021autoencoder} proposed a regularization method that molds the latent space into a smooth, locally convex manifold consistent with training images. \cite{pinkney2020resolution} presents a method for interpolating between generative models of the StyleGAN architecture in a resolution-dependent manner. However, these approaches fail to capture the substantial alterations in spatial correlation present in 3D image data.

\subsection{Explanation Supervision}
Incorporating human knowledge into explainable models has been a central focus of research in natural language and tabular data, utilizing methods like attribution and feature regularization~\cite{adadi2018peeking}. XAI-Class~\cite{arrieta2020explainable} utilize highlighted words from the input as additional signals for training the Transformer model. Commonsenseqa~\cite{talmor2018commonsenseqa} proposes to train language models in a multi-task manner, supervised by both labels and rationales~\cite{zhang2024elad}. Recently, there has been a growing recognition of the value of visual explanations. A leading method to achieving this involves the use of saliency maps, which identify the input features most influential to a model's predictions~\cite{montavon2019layer, selvaraju2017grad}. The HAICS framework~\cite{shen2021human} represents a notable advancement in image classification, utilizing human-generated scribble annotations as the explanation supervision signal. RES~\cite{gao2022res} introduced an innovative objective designed to accommodate the inaccurate, incomplete, and inconsistent nature of human annotations. MAGI~\cite{zhang2023magi} proposes a generative model to solve the multiple noisy and incomplete annotations for supervision. Despite this advancement, research on applying explanation supervision to 3D data remains under-explored.

\section{Problem Formulation} \label{sec:problem_formulation}

This section presents problem formulation regarding explanation supervision in the context of image classification. We first define the general paradigm of explanation supervision, and then explore the extension of this paradigm from 2D to 3D data, introducing unique challenges posed by the additional dimensionality. Problem formulation for 3D explanation supervision is presented as follows:



Given a set of inputs \(X = \{x_{i}\}_{i=1}^N\) with class labels \(Y = \{y_{i}\}_{i=1}^N\), and their corresponding human explanation annotation \(M = \{m_{i}\}_{i=1}^N\), where $N$ denotes the training sample size, the model aims to learn the mapping function $f(\cdot)$ for each input image $x_{i}$ to its class label as $f: x \to y$ and provide a model explanation via an explainer $g(\cdot)$ as $g: (f, \langle x,y\rangle) \to m$. The general paradigm of \textit{explanation supervision} can be formulated as the objective function below:

\begin{equation}
    \min \sum\nolimits_{i=1}^{N} (\underbrace{\mathcal{L}_{\text{Pred}}(f(x_{i}, y_{i})}_{\text{prediction loss}} + \lambda \underbrace{\mathcal{L}_{\text{Exp}}(g(f,\langle x_{i}, y_{i}\rangle), m_{i})}_{\text{explanation loss}})
    \label{eq:general_explanation_supervision}
\end{equation}
where the first term measures the prediction loss of the model's predicted class labels, the second term measures the explanation loss of the model-generated explanation, and $\lambda$ is a hyper-parameter used to balance the two loss terms. Here, $\mathcal{L}_{\text{Pred}}$ represents a common prediction loss (e.g., cross-entropy loss), while $\mathcal{L}_{\text{Exp}}$ is tailored to the characteristics of individual datasets.

In our study, we want to extend the above  \textit{explanation supervision} paradigm to 3D data as $X = \{x_{i} \in \mathbb{R}^{C \times D \times H \times W}\}_{i=1}^N$ with class labels $Y = \{y_{i}\}_{i=1}^N$, and their corresponding binary annotations $M = \{m_{i} \in \mathbb{R}^{D \times H \times W}\}_{i=1}^N$, where $N$ denotes the training sample size, $C$ denotes the number of channels, $D$ denotes depth, $H$ denotes height, and $W$ denotes width. However, various challenges arise when extending this paradigm to 3D data:
\textbf{1) Spatial correlation changed.} Spatial correlation has changed significantly from 2D to 3D image data. The complexity of 3D data requires adjustments to capture features and correlations in the third dimension, potentially causing misalignment between data patterns and the paradigm's capacity.
\textbf{2) Absence of direct 3D labeling.} The absence of direct 3D labeling poses a challenge because human labeling is initially 2D. Manual labeling of volumetric data is costly and leads to sparse labeling, especially in the depth dimension. Limited training samples in domains such as medical imaging exacerbate the challenge of effective generalization.
\textbf{3) Uncertainty varies across different parts of the explanation.} Medical imaging data are shaped by anatomical structures and lesions, which exhibit a diverse range of shapes. The inconsistent human labeling further contributes to varying distances between consecutive slices along the depth dimension, introducing dynamic uncertainty. Addressing these uncertainties poses a challenge to the model's capacity to provide consistent explanations across different regions of 3D data.



\section{Proposed Framework} \label{sec:method}
This section describes the proposed DUE framework in detail. We begin with an overview of the framework and then describe its key components.

\begin{figure}[h]
  \centering
  \includegraphics[width=\linewidth]{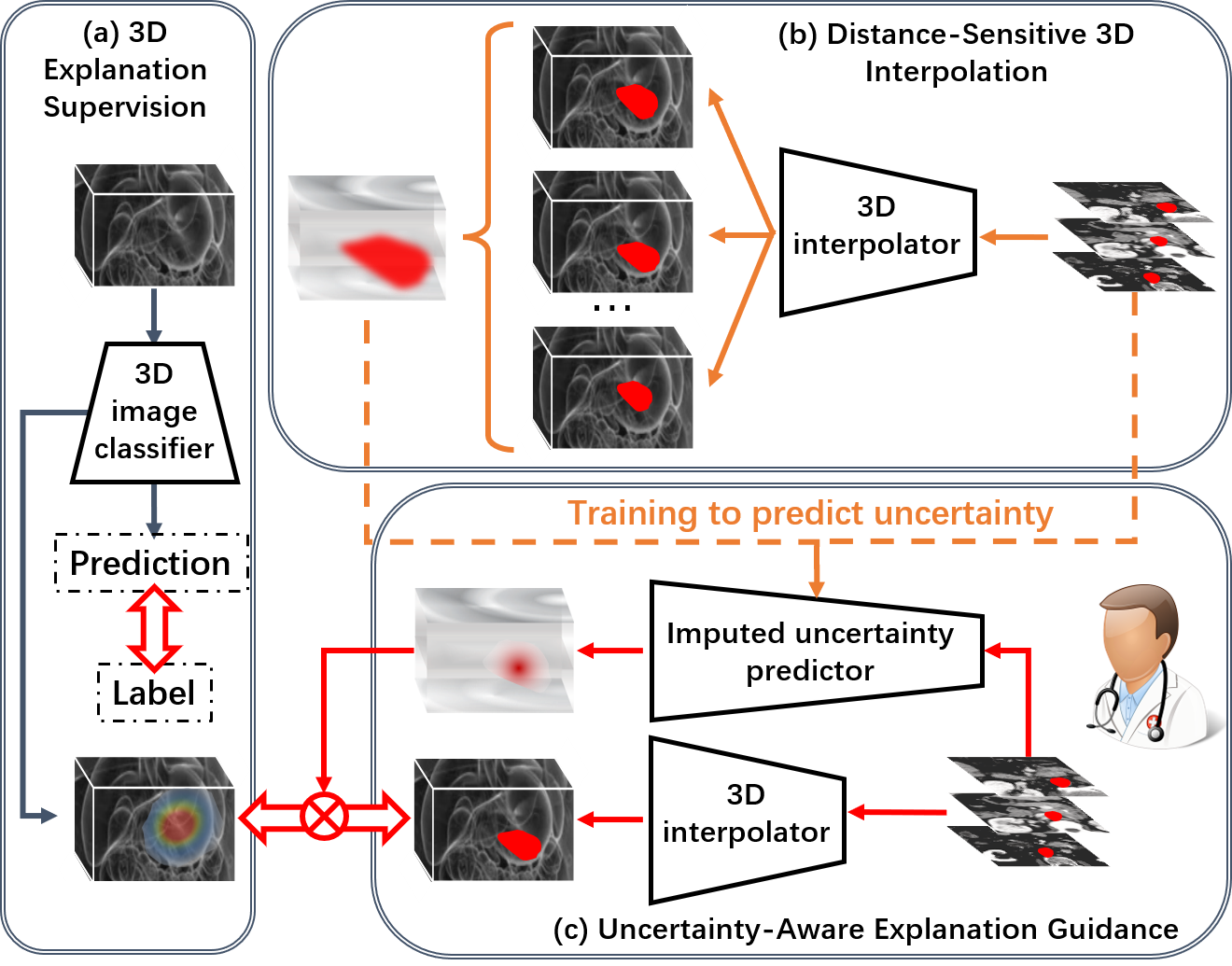} 
  \caption{Overview of the DUE framework: (a) presents the 3D explanation supervision, (b) demonstrates the Distance-Sensitive 3D interpolation, and (c) illustrates the Uncertain-Aware Explanation Guidance.}
  \label{fig:framework}
\end{figure}

\subsection{Framework Overview}
The proposed DUE framework, as shown in Figure~\ref{fig:framework}, consists of three modules: \textit{3D explanation supervision}, \textit{distance-sensitive 3D interpolation}, and \textit{uncertain-aware explanation guidance}. 

The 3D explanation supervision module aims to enhance both the predictability and explainability of models by introducing 3D explanation supervision. This approach incorporates 3D explanation annotations as additional supervision, alongside traditional prediction label supervision, as depicted in Figure~\ref{fig:framework}(a). While ideal explanation guidance requires comprehensive and precise explanation annotations specifying areas of focus at the pixel (or voxel) level, the sparsity of annotation slices serves as incomplete guides, limiting the effectiveness of explanation supervision. In this scenario, addressing missing slices becomes crucial for effective 3D visual explanation guidance. Traditional interpolation methods such as linear interpolation~\cite{maeland1988comparison} can fill in missing slices, but may introduce bias since they ignore the proportional relationship between uncertainty and the distance from conditional slices. Furthermore, they only provide deterministic predictions, neglecting randomness observed in real-world scenarios~\cite{babaeizadeh2017stochastic}, making them less suitable compensatory approaches.

To address this issue, we propose a distance-sensitive 3D interpolation module, which comprises a 3D interpolator to better account for the proportional relationship between uncertainty and the distance from the conditional slices when interpolating missing slices. 
The 3D interpolator is extended from a conditional diffusion model, detailed in Section~\ref{subsec:Diffusion}.

Subsequently, we introduce an uncertain-aware explanation guidance module to estimate the uncertainty of the interpolated slices and determine the weights assigned to the interpolated slices based on their associated uncertainty. Specifically, we utilize an imputed uncertainty predictor to accelerate the uncertainty estimation procedure via neural processes, which will be elaborated in Section~\ref{subsec:UQ}. These uncertainties are translated into weights for each voxel contributing to the final 3D explanation, tuning their influence as uncertain-aware explanation guidance, as shown in Figure~\ref{fig:framework}(c). Subsequently, the above two modules are synergistically incorporated into the 3D explanation supervision framework, enabling effective handling of challenges arise from the absence of direct 3D annotations. 

Based on the above statement, the proposed DUE framework achieves 3D explanation supervision through the integration of 3D interpolation uncertainty prediction and uncertainty-aware explanation annotation guidance, as depicted in Figure~\ref{fig:framework}. Formally, the overall objective of the DUE framework is expressed as follows:
\begin{flalign}
\begin{aligned}
    \min_{g, f} & \sum\nolimits_{i=1}^{N_{all}} \mathcal{L}_{\text{Pred}} (f(x_{i}), y_{i}) + \\
    \lambda & \sum\nolimits_{i=1}^{N_{pos}} \mathcal{L}_{\text{Exp}} \left(g(f, \langle x_{i}, y_{i} \rangle), G_{\text{imp}}(m_{i}) \cdot g_{\text{interp}}(m_{i}) \right),
\end{aligned}
\label{eq:due_objective}
\end{flalign}
where prediction loss is computed for all samples, while explanation loss is computed only for samples with manual labels (i.e., positive samples). The function $G_{\text{imp}}(\cdot)$ represents the imputed uncertainty predictor, tasked with adjusting the influence of annotations on the explanation loss based on their imputed uncertainty. Additionally, the 3D interpolator is denoted as $g_{\text{interp}}(\cdot)$.


\subsection{Distance-Sensitive 3D Interpolation}
\label{subsec:Diffusion}
To enhance the proportional relationship between uncertainty and distance from conditional slices, we involve a 3D interpolation method that extends a conditional diffusion model to perform distance-sensitive interpolation for the missing slices~\cite{voleti2022MCVD}.
When interpolating missing annotation slices, consider a set \(\mathcal{A} = \{A_i \in \mathbb{R}^{H \times W}\}_{i=1}^N\) representing the annotation slices of an incomplete annotation, where $N$ is the total number of slices. Each annotation slice $A_i$ contains the contour line of the region of interest. Let $D = \{d_1,  \in \mathbb{R}\}_{i=1}^{N-1}$ be the corresponding set of distances between adjacent annotation slices, where $d_i$ denotes the distance between slices $A_i$ and $A_{i+1}$. The goal is to interpolate missing slices \(\mathcal{A}_{\text{missing}} = \{A_{\text{missing},i}\in \mathbb{R}^{H \times W}\}_{i=1}^{M}\) within this set based on the set of existing slices \(\mathcal{A}\), where $M$ is the number of missing slices.

Given the considerable time consumption and computational complexity associated with 3D interpolation, we adopt a chunking approach to interpolate the entire annotation. Initially, the annotation is segmented into multiple blocks, each comprising two slices. Subsequently, interpolation is applied to fill in the missing slices within each interval, with conditioning on the two slices. The interpolation process adopts an autoregressive approach, allowing each new interpolation conditioned on the preceding interpolations. This generates \(j < d_i\) slices iteratively until a cumulative total of \(d_i\) slices is attained. By amalgamating all the missing slices within each interval \(\sum_{i=1}^{N-1} d_i\), the total number of slices \(M\) is obtained. 

\begin{figure}[h]
  \centering
  \includegraphics[width=\linewidth]{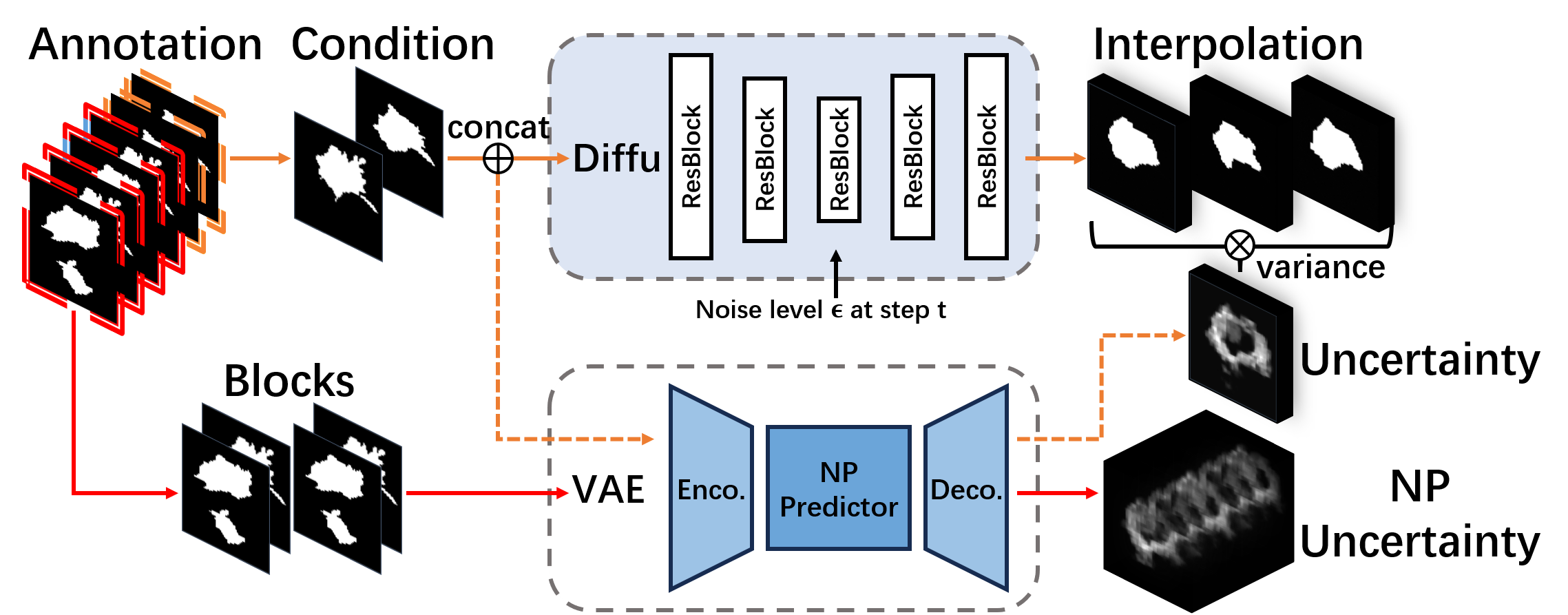} 
  \caption{Overview of the \textit{imputed uncertainty predictor} training: A diffusion model is first trained for interpolation and uncertainty generation (solid orange line). Then, a Neural Processes (NP)-based VAE is trained to impute uncertainty for NP representations (dashed orange line). The red line represents the deployment path.}
  \label{fig:uq}
\end{figure}

\noindent \textit{\textbf{Conditional diffusion for slice interpolation.}}
Based on the above discussion, consider an annotation block comprising two annotation slices, denoted as $A_i$ and $A_{i+1}$, utilized as conditions. The diffusion model is tasked with interpolating each intermediate slice $A_j$, where $j$ ranges from 1 to $d_i$, as shown in Figure~\ref{fig:uq}. This interpolation process can be formulated as follows:
\begin{flalign}
    \begin{aligned}
    & \mathbb{E}_{t, [A_i, A_j, A_{i+1}] \sim p_{\text{annotation}}, \epsilon \sim \mathcal{N}(0, I), (m_p, m_f) \sim \text{B}(p_{\text{mask}})} \\
    & \quad \left[ ||\epsilon - \epsilon_\theta \left( \sqrt {\bar{\alpha}_t} A_j + \sqrt{1 - \bar{\alpha_t}} \epsilon \,\middle|\, m_p A_i, m_f A_{i+1}, t \right) ||^2 \right],
    \end{aligned}
\label{eq:diffusion_loss_function}
\end{flalign}
where $m_p$ and $m_f$ are all-zero masks independently sampled from the Bernoulli distribution $\mathcal{B}$ with a probability of $p_{\text{mask}} = 1/2$. The term $p_{\text{annotation}}$ represents the distribution of annotation slices. The function $\epsilon_\theta(A_{j,t}|t)$ estimates $\epsilon$ through a time-conditional neural network parameterized by $\theta$, and $\bar{\alpha}_t$ is a parameter that regulates the balance between the contribution of the interpolated slice $A_j$ and the noise term $\epsilon$ at step $t$ in the \textit{reverse process}~\cite{ho2020denoising}. 


\subsection{Uncertainty-Aware Explanation Guidance}
\label{subsec:UQ}
After obtaining the interpolated slices and the 3D interpolator, we introduce the uncertain-aware explanation guidance module for generating associated uncertainties. Specifically, we involve an imputed uncertainty predictor to dynamically estimate voxel uncertainty levels, represented as \(U=\{u^{(i)} \in \mathbb{R}^{H \times W}\}_{i=1}^M\), for intervals within the annotation slices by utilizing spatial correlation and the distribution of annotation features, conditioned on two ground truth slices. After determining the voxel uncertainties of interpolated 3D annotations, we translate the uncertainties into weights, then multiply weights by the interpolated annotations to obtain the final, complete explanatory annotation. 

Initially, we use the diffusion model described in Section~\ref{subsec:Diffusion} to iteratively perform the interpolation process and compute variance, as illustrated in Algorithm~\ref{alg:gen_var}. 
The variance is used as an approximation to uncertainty. The iterative interpolation process and the computation of variance are carried out from lines 4 to 8. The interpolation procedure is detailed from lines 10 to 15, while the function for computing variance spans from lines 16 to 21. This approach enables a quantitative assessment of uncertainty, offering insights into the randomness inherent in the interpolation results. These estimations are then used to weight the interpolated annotation slices, tuning their influence as explanation guides. 

\begin{algorithm}
\caption{Algorithm for Variance Generation}
\label{alg:gen_var}

\begin{algorithmic}[1]
    \State \textbf{Input:} Training data $D$, number of iterations $T$
    \State \textbf{Output:} Variance $V$
    \State Train a diffusion model on $D$.
    \For{$t = 1$ to $T$}
        \State Interpolate segments of the annotation using the diffusion model: $A_t \gets$ \Call{3DInterpolation}{$D$}.
        \State Aggregate interpolated segments: $A \gets A \cup A_t$.
    \EndFor
    \State Compute the variance of all outcomes: $V \gets$ \Call{ComputeVariance}{$A$}.
    \State \textbf{return} $V$.
    
    \Function{3DInterpolation}{$D$}
        \State \textbf{Input:} Training data $D$.
        \State \textbf{Output:} Interpolated annotation segments $A_t$.
        \State Interpolate $D$ using the diffusion model, following the specified loss function.
        \State \textbf{return} Interpolated segments $A_t$.
    \EndFunction
    
    \Function{ComputeVariance}{$A$}
        \State \textbf{Input:} Aggregate of interpolated annotation segments $A$.
        \State \textbf{Output:} Variance $V$.
        \State Calculate the variance of outcomes across all $A$.
        \State \textbf{return} Calculated variance $V$.
    \EndFunction
\end{algorithmic}
\end{algorithm}

As in the above statements, we use diffusion models for interpolation and generate the associated uncertainties. However, generating variance with diffusion models is slow and inherently unstable. To address this, we substitute the diffusion model with a Variational Autoencoder (VAE) based on Neural Processes (NP)~\cite{ye2023unified} for more stable and swift variance generation. 
The VAE model enables a continuous mapping of uncertainty and promptly generates it, alleviating concerns about deployment speed and stability associated with the diffusion model. Additionally, neural processes accommodate varying spacing between annotation slices. To emphasize the purpose of the VAE, we name it as \textit{imputed uncertainty predictor} in Figure~\ref{fig:framework}. 

The learning process of the imputed uncertainty predictor comprises two stages, as shown in Figure~\ref{fig:uq}. Initially, the diffusion model is trained for interpolation and uncertainty generation (depicted by the solid orange line). Subsequently, we utilize the same condition slices from the diffusion model as conditions for the VAE, with the variances generated by the diffusion model serving as the targets. The VAE is trained to expand the representation of uncertainty for varying spacing (illustrated by the dashed orange line). The reason for this two-stage approach is the complexity associated with training diffusion models and VAEs, which makes simultaneous training a challenging task. During deployment, the VAE is employed directly, as indicated by the red line. Upon obtaining the uncertainty, it is converted into weights using a simple mapping: applying a flipped $Sigmoid$ function to the uncertainty, followed by min-max normalization. After the training process, we can generate the NP uncertainty of a given interpolated 3D annotation with the VAE model.

\section{Experiments} \label{sec:experiment}

We present a comprehensive analysis of the experimental results of our proposed framework, focusing on two tasks: pancreatic tumor classification and lung nodule classification. We first introduce our experimental settings, including tasks and datasets, evaluation metrics, and comparative methods. Subsequently, we conduct an extensive quantitative assessment of the model's predictions and explanations and simulate real-world scenarios by restricting training samples. Additionally, we conduct an ablation study, a qualitative assessment featuring case studies, and a sensitivity analysis to provide further insights.

\subsection{Experimental Settings}

\noindent \textit{\textbf{Pancreatic tumor classification:}}
We obtained negative samples (i.e., normal samples) from the Pancreas-CT dataset\footnote{Available at: https://wiki.cancerimagingarchive.net/display/Public/Pancreas-CT}~\cite{roth2015deeporgan} and positive samples (i.e., abnormal samples) from the Medical Segmentation Decathlon dataset\footnote{Available at: http://medicaldecathlon.com/}, resulting in a dataset of 281 CT scans with tumors and 80 CT scans without tumors. The pancreas region was extracted based on doctors' annotations while retaining the presence of tumors. We kept the original 3D modality for the samples, and extracted the middle slice along the depth dimension for 2D comparative methods, resulting in $128 \times 128 \times 64$ image blocks and $128 \times 128$ image slices, respectively. We split the dataset into 30\% for training and validation, and 70\% for testing, maintaining a balanced data distribution for training while keeping the original ratio for validation and test sets. In our experiments, we only utilized 20 samples during training to simulate a real-world scenario where manual labels are strictly limited.

\noindent \textit{\textbf{Lung nodule classification:}} 
We obtained both positive samples (i.e., samples with nodules) and negative samples (i.e., non-nodule samples) from the LIDC-IDRI dataset\footnote{Available at: https://wiki.cancerimagingarchive.net/pages/viewpage.action?pageId=1966254}~\cite{armato2011lung}. This dataset includes CT scans collected from 1010 patients, accompanied by annotations provided by four experienced radiologists. Employing a standard 50\% consensus consolidation of these annotations, we identified nodule regions as positive samples and the surrounding areas as negative samples, resulting in 2625 positive samples and 68,160 negative samples. We retained the original 3D modality for the samples and extracted the middle slice along the depth dimension for 2D comparative methods. This yielded image blocks of size $128 \times 128 \times 64$ and image slices of size $128 \times 128$, respectively. The dataset was first split at patient level, allocating 10\% for training, 30\% for validation, and 60\% for testing. We ensured a balanced distribution of the training data while preserving the original distribution of validation and test sets. To simulate real-world scenarios, we conducted experiments using 20, 50, and 100 training samples.

\begin{table*}[htbp]
    \centering
    \caption{The experimental results comparing model prediction and generated explanations to various methods for pancreatic tumor and lung nodule classification tasks. Optimal outcomes for each task are highlighted in bold.}
    \label{result}
    \begin{tabular}{l|c|cc|cccc}
        \toprule
        Dataset & Method & ROC-AUC (↑) & PR-AUC (↑) & IoU (↑) & Precision (↑) & Recall (↑) & F1 (↑) \\
        \midrule
    \multirow{6}{*}{Pancreas} & HAICS & $76.37 \pm 10.26$ & $89.35 \pm 6.09$ & $33.63 \pm 3.95$ & $99.83 \pm 0.29$ & $40.29 \pm 5.20$ & $54.83 \pm 4.40$ \\
    & GRADIA & $76.41 \pm 11.62$ & $90.35 \pm 6.52$ & $36.40 \pm 5.59$ & $99.83 \pm 0.29$ & $45.13 \pm 7.18$ & $59.84 \pm 6.95$ \\
    & RES & $77.90 \pm 11.02$ & $90.57 \pm 6.52$ & $35.47 \pm 8.36$ & $99.66 \pm 0.59$ & $42.68 \pm 11.31$ & $57.23 \pm 11.77$ \\
    & Baseline & 96.83 $\pm$ 4.22 & 99.14 $\pm$ 1.12 & 38.81 $\pm$ 16.85 & 99.83 $\pm$ 0.29 & 49.11 $\pm$ 21.55 & 62.60 $\pm$ 20.27 \\
    & Baseline$^{+}$ & 96.51 $\pm$ 5.42 & 99.17 $\pm$ 1.26 & 36.88 $\pm$ 5.39 & \textbf{100 $\pm$ 0} & 42.40 $\pm$ 6.75 & 57.33 $\pm$ 6.31 \\
    & \textbf{DUE (proposed)} & \textbf{99.58 $\pm$ 0.24} & \textbf{99.88 $\pm$ 0.07} & \textbf{51.20 $\pm$ 2.43} & \textbf{100 $\pm$ 0} & \textbf{63.43 $\pm$ 4.19} & \textbf{75.66 $\pm$ 2.76} \\
    \midrule
    \multirow{6}{*}{LIDC} & HAICS & $62.29 \pm 14.20$ & $8.25 \pm 6.63$ & $26.77 \pm 4.49$ & $99.74 \pm 0.45$ & $47.60 \pm 9.36$ & $62.22 \pm 9.27$ \\
    & GRADIA & $58.37 \pm 12.05$ & $7.72 \pm 3.21$ & $29.46 \pm 5.30$ & \textbf{100 $\pm$ 0} & $52.69 \pm 8.18$ & $67.27 \pm 7.31$ \\
    & RES & $65.84 \pm 17.37$ & $14.59 \pm 16.28$ & $28.53 \pm 8.69$ & $99.48 \pm 0.45$ & $50.05 \pm 16.83$ & $64.00 \pm 16.29$ \\
    & Baseline & 97.60 $\pm$ 0.39 & 81.72 $\pm$ 1.94 & 14.30 $\pm$ 6.58 & 94.32 $\pm$ 1.79 & 30.10 $\pm$ 13.76 & 40.42 $\pm$ 14.38 \\
    & Baseline$^{+}$ & 96.55 $\pm$ 1.76 & 79.81 $\pm$ 4.35 & 31.47 $\pm$ 0.41 & 97.67 $\pm$ 2.33 & 43.84 $\pm$ 5.12 & 56.99 $\pm$ 4.18 \\
    & \textbf{DUE (proposed)} & \textbf{98.58 $\pm$ 0.38} & \textbf{87.82 $\pm$ 1.25} & \textbf{33.28 $\pm$ 2.69} & 90.18 $\pm$ 3.82 & \textbf{64.99 $\pm$ 4.39} & \textbf{67.66 $\pm$ 4.25} \\
    \bottomrule
    \end{tabular}
\end{table*}

\begin{figure*}[h]
  \centering
  \includegraphics[width=0.9\linewidth]{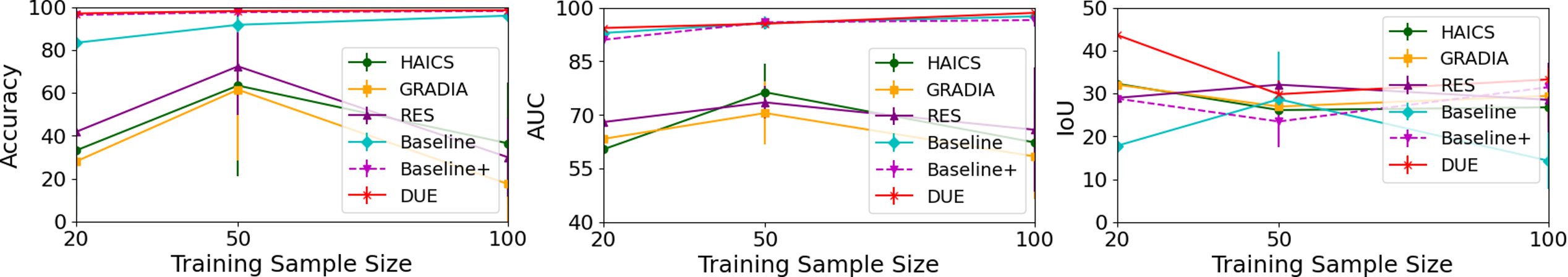} 
  \caption{Model performance under varying training sample sizes on the lung nodule classification dataset. (Left) Comparison of test prediction accuracy. (Middle) Comparison of test prediction ROC-AUC. (Right) Comparison of test IoU score.}
  \label{fig:sample_size}
\end{figure*}

\begin{figure*}[h]
  \centering
  \includegraphics[width=0.9\linewidth]{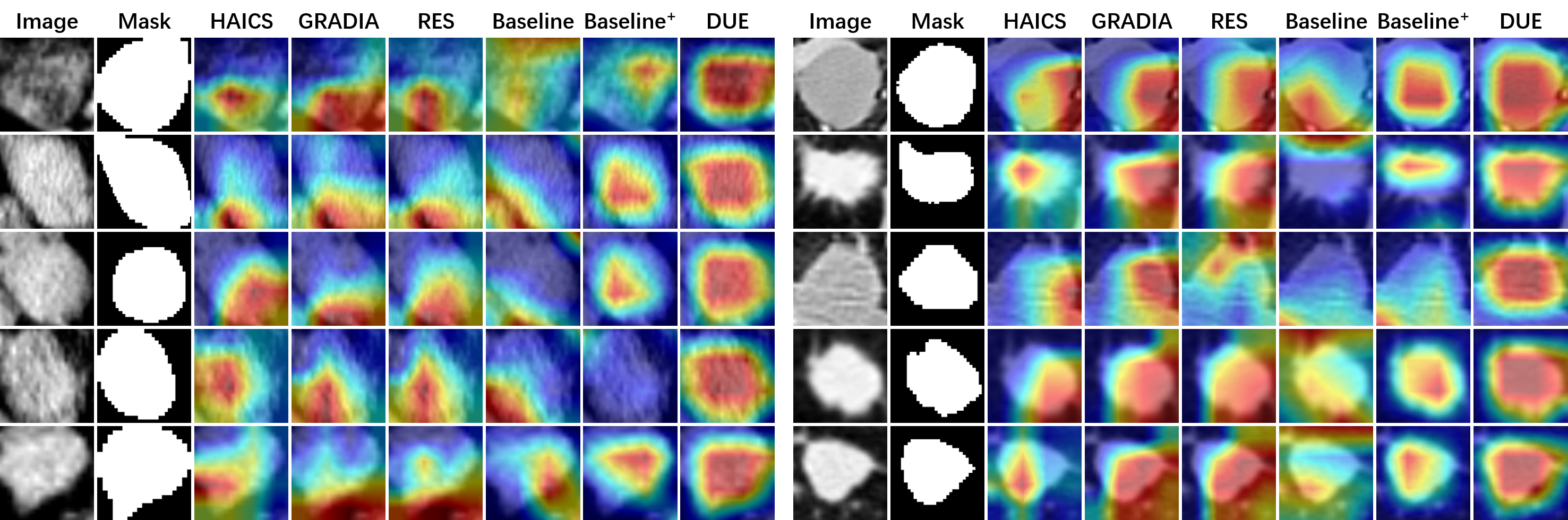} 
  \caption{Visualizations display explanations for pancreatic tumor classification (left) and lung nodule classification (right). Human annotations are presented in the Mask columns, while model-generated explanations are depicted using heatmaps overlaid on the original images, highlighting regions of greater importance with warmer color intensities.}
  \label{fig:heatmap}
\end{figure*}

\noindent \textit{\textbf{Evaluation metrics:}}
In assessing the model's performance, we consider both its predictability and explainability. To evaluate its predictive capabilities, common metrics such as prediction accuracy and the Area Under the Curve of the Receiver Operating Characteristic (ROC-AUC) curves are employed. Due to the pronounced imbalance in the test sets (with a positive-to-negative ratio of approximately 1:26 for LIDC), we additionally employ the Area Under the Curve of the Precision-Recall (PR-AUC) curves as a metric~\cite{davis2006relationship}. To assess the quality of the model's explanations, we compare its generated explanations with human annotations. Specifically, we utilize the Intersection over Union (IoU) score, as introduced in ~\cite{bau2017network}. 
This score is derived from the bit-wise intersection and union operations between the human explanations and the binarized model-generated explanations, providing a measure of overlap between the two inputs. 
Additionally, we compute pixel-wise precision, recall, and F1 score, offering a comprehensive evaluation of the model-generated explanations. 

\noindent \textit{\textbf{Comparative methods:}}
We conduct a comparative analysis by evaluating our proposed method against three existing explanation supervision methods: HAICS~\cite{shen2021human}, GRADIA~\cite{gao2022aligning}, and RES~\cite{gao2022res}. Additionally, we include the baseline, which is a 3D model trained solely with the prediction loss. Furthermore, as part of an ablation study, we assess two variants of our proposed method, namely Baseline$^{+}$ and the standard DUE. 

\begin{itemize}

\item \textbf{HAICS~\cite{shen2021human}:} This framework employs explanation supervision to train a 2D model, utilizing Binary Cross Entropy (BCE) loss to minimize the discrepancy between the model-generated explanations and provided explanation annotations.

\item \textbf{GRADIA~\cite{gao2022aligning}:} This framework employs explanation supervision to train a 2D model, utilizing L1 loss to minimize the discrepancy between the model-generated explanations and provided explanation annotations.

\item \textbf{RES~\cite{gao2022res}:} This framework trains a 2D model with robust explanation supervision, utilizing imputation to bidirectionally minimize the distance between the model's explanations and the provided explanation annotations.

\item \textbf{Baseline:} The backbone model of our method, which is a bare 3D model trained solely with the prediction loss.

\item \textbf{Baseline$^{+}$:} A naive variant of our method, which trains a 3D model using explanation supervision that directly minimizes the distance between the model's explanations and the provided explanation annotations.

\item \textbf{DUE:} The standard variant of our method, which trains a 3D model using explanation supervision with uncertainty-aware explanation guidance to minimize the distance between the model's explanations and the provided explanation annotations.

\end{itemize}

\noindent \textit{\textbf{Implementation details:}}
The backbone model utilized is a 3D ResNet18~\cite{hara3dcnns}for all 3D methods. For the 2D methods, ResNet18~\cite{he2016deep} is employed with customization. This customization involves adjusting the first convolutional layer to possess a kernel size of (7, 7), a stride of (2, 2), and padding of (3, 3). This modification aligns the feature map's view with that of the 3D models, thus mitigating resolution discrepancies and enabling a fair comparison. For RES~\cite{gao2022res}, we use the Gaussian imputation (i.e., RES-G) and set $\alpha$ to 0.001. All explanation supervision methods employ an attention weight $\lambda$ of 1.
All models undergo training for 50 epochs using the Adam optimizer~\cite{kingma2014adam} with a learning rate set at 0.001. Model explanations are generated via Grad-CAM~\cite{selvaraju2017grad} and binarized using a threshold of 0.5. 

\subsection{Performance}

Table ~\ref{result} shows the model prediction and generated explanation performance for the pancreatic tumor and lung nodule classification datasets. The results are obtained from 5 individual runs. The best results for each dataset are highlighted in bold. Overall, our proposed framework DUE outperformed all other comparison methods in both prediction performance and explainability on both datasets. In addition, the huge difference in predictive abilities between 3D and 2D models is due to the underutilized information in 3D data, which further demonstrates the significance of extending the primitive explanation supervision paradigm to 3D models.

For the pancreatic tumor classification task, our proposed DUE consistently yields the best performance on all metrics. Specifically, DUE achieved the highest ROC-AUC and PR-AUC, outperforming the baseline and other comparison methods by 2.75\%-23.21\%, and 0.71\%-10.53\% respectively for the predictive capability. The improvements are even more on the model explainability in terms of IoU, recall, and F1 scores, where DUE exceeded baseline and other comparison methods by 12.39\%-17.57\%, 14.32\%-23.14\%, and 13.06\%-20.83\% respectively. Additionally, precision is at a perfect 100\% for DUE.

Furthermore, for the lung nodule classification task, the DUE shows stronger predictive capabilities, where the ROC-AUC and PR-AUC scores are 0.98\%-40.21\%, and 6.1\%-80.1\% higher for the DUE compared to baseline and other comparison methods. While precision has a decrease of 4.14\% compared to the baseline model, this is offset by a significant improvement in recall with an increase of 34.89\%. This balance between precision and recall is reflected in the IoU, with an improvement of 18.98\% over the baseline model, as well as an overall increase in the F1 score by 27.24\%, indicating enhancing model explainability robustly.

Subsequently, we investigate the performance of the DUE framework to strengthen the generalization capabilities under various training sample sizes. We study three training sample sizes of 20, 50, and 100 using the lung nodule classification dataset. As depicted in Figure ~\ref{fig:sample_size}, we present the results of the prediction accuracy, AUC, and IoU score of each method concerning the training sample size. Each data point represents the mean values of five independent runs and the corresponding error bar stands for the standard deviation. In general, our DUE framework outperformed all other comparison methods, especially in the predictive capabilities, demonstrating the effectiveness of our proposed framework. Specifically, DUE can improve the prediction accuracy and AUC by 40\% and 30\% respectively on average against other comparison methods. Interestingly, DUE performs the best in the IoU score when the sample size is as small as 20. As the training sample size increases, the IoU score decreases and then stabilizes while still outperforming other comparison methods, indicating a transition from potential overfitting to better generalization. 

\subsection{Qualitative Analysis of Model Explanation}
\label{subsec:QualitativeExp}

Here, we present a case study examining the comparison of model-generated explanations for pancreatic tumor and lung nodule classification datasets, as depicted in Figure~\ref{fig:heatmap}. The model-generated explanations are showcased through heatmaps overlaid on the original image samples, with increased emphasis on areas exhibiting warmer colors to denote higher importance.

\noindent \textit{\textbf{Pancreatic tumor classification:}}
In the context of pancreatic tumor classification, illustrated in the left portion of Figure~\ref{fig:heatmap}, we chose five samples of model-generated explanations from all models. The visualization reveals that explanations produced by models employing the proposed DUE framework exhibit superior performance in both accuracy and alignment with human annotations compared to the comparative methods. Notably, the explanations generated by the baseline model largely fail to concentrate on the annotated region, indicating its inherent lack of explainability. The 2D methods exhibit a recurring focus on the lower region, consistently focusing on the lower region, while the ground truth is distributed in the middle of the view. Generally, Baseline${+}$ manages to anchor on the correct region but with a narrow scope deviating from the central tumor area. DUE achieves optimal visualization by consistently focusing on the central area, with margin adjustments to accurately encompass the tumor region.

\begin{figure}[h]
  \centering
  \includegraphics[width=\linewidth]{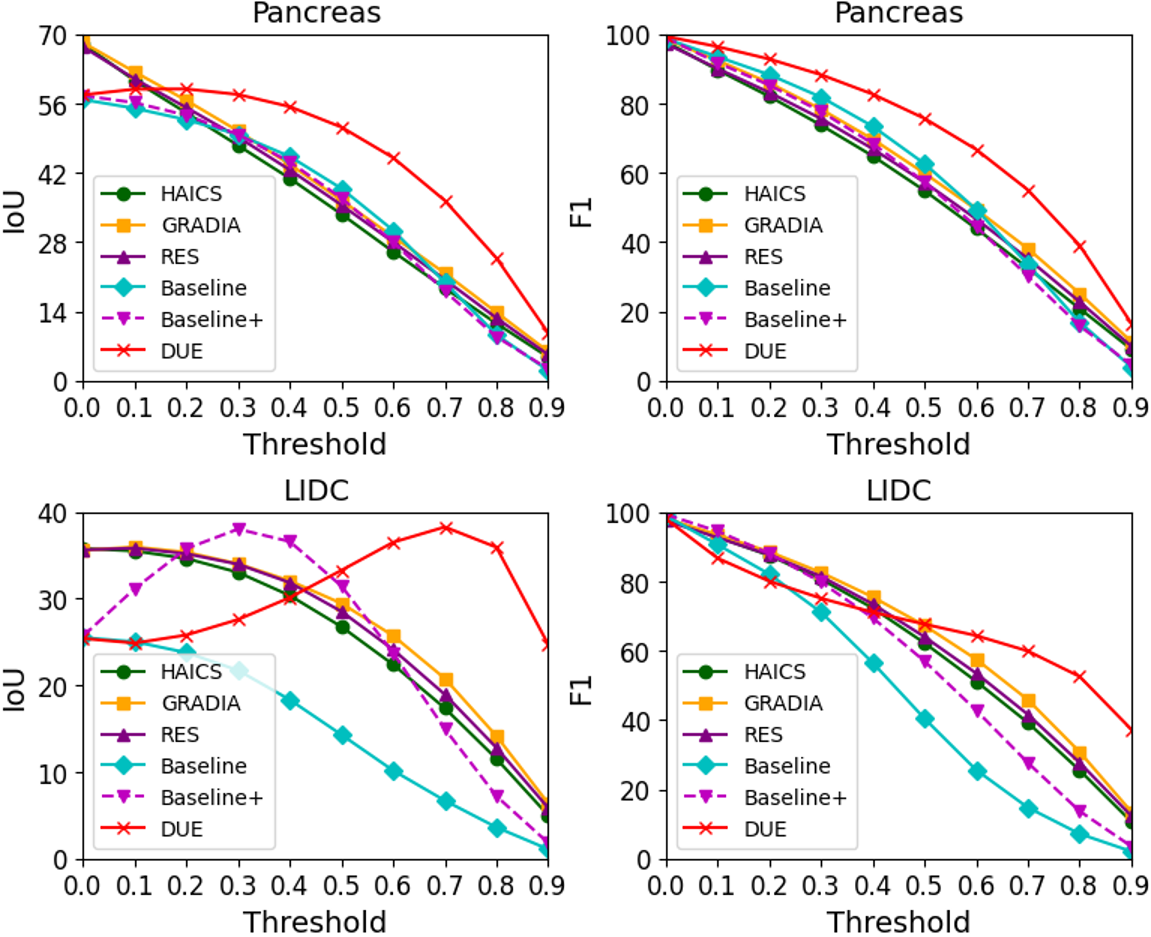} 
  \caption{IoU and F1 scores for model explanations at different thresholds on pancreatic tumor and lung nodule classification datasets.}
  \label{fig:All_Exp}
\end{figure}

\noindent \textit{\textbf{Lung nodule classification:}} 
In the right segment of Figure~\ref{fig:heatmap}, five selected samples of model-generated explanations across all models are presented for lung nodule classification. The contours of the lung nodules exhibit greater variability, with DUE consistently delivering the most effective visualization results. Specifically, the baseline model continues to favor areas outside the annotated region, indicating a questionable basis for its predictions. The 2D methods consistently focus on the lower right part, capturing a significant portion of the ground truth but also encompassing large non-relevant areas. Baseline$^{+}$ achieves precise concentration; however, it remains incomplete and repeats the same bias observed in the baseline for row 3. In contrast, DUE tends to encompass the annotated region as extensively as possible while also adjusting its margins to prevent overreaching. Furthermore, DUE exhibits strong confidence in its attention area, as indicated by its heatmap being predominantly red with a narrow margin of yellow.

\subsection{Quantitative Analysis of Model Explanation}

\begin{figure}[h]
  \centering
  \includegraphics[width=0.8\linewidth]{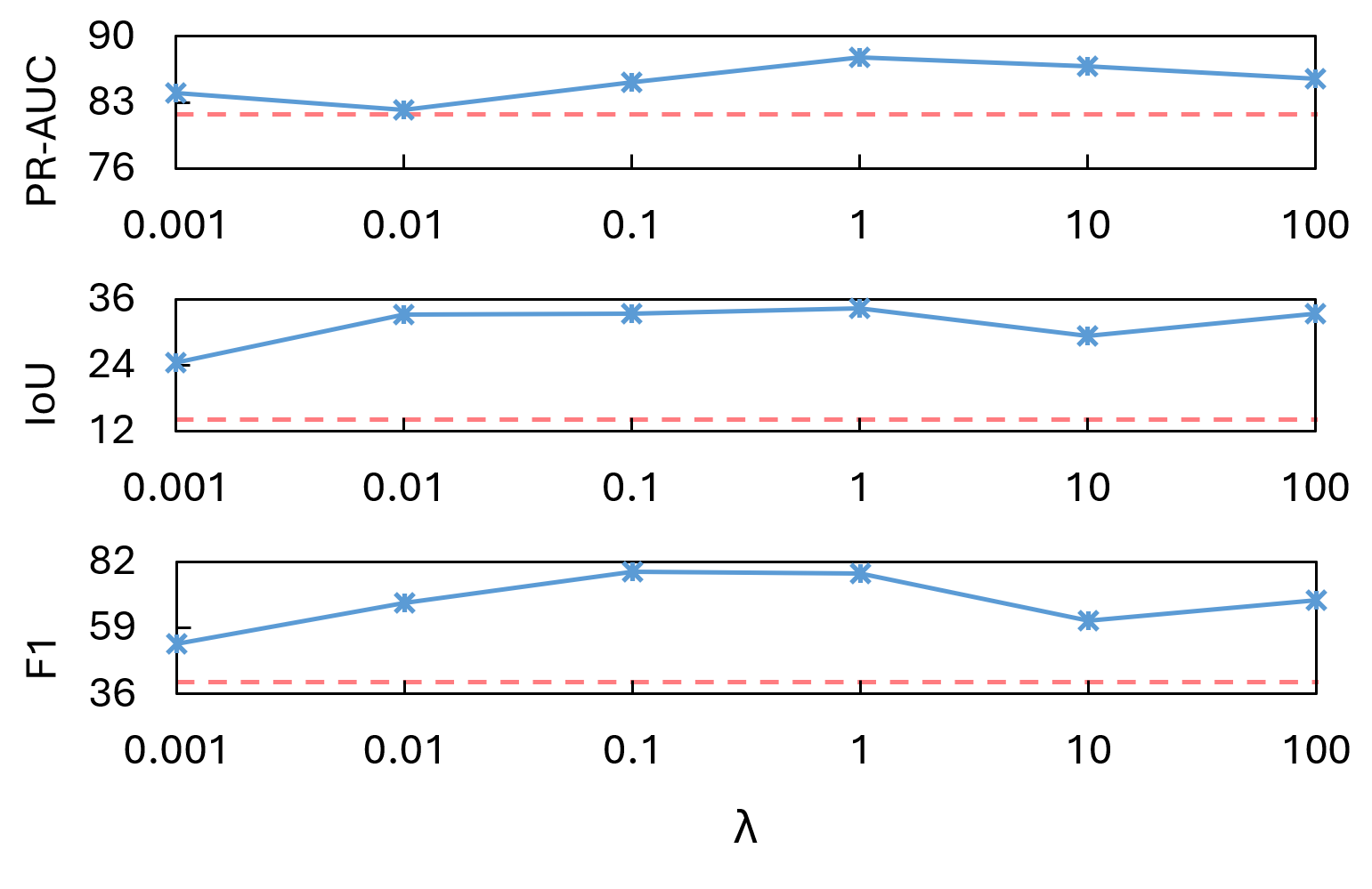} 
  \vspace{-0.3cm}
  \caption{The sensitivity study of \(\lambda\) in our framework, DUE, on the lung nodule classification dataset. The red dashed line denotes the baseline model's performance.}
  \label{fig:sensitivity}
\end{figure}

In addition to the qualitative analysis of model explanations presented in Section~\ref{subsec:QualitativeExp}, which is conducted with a limited number of samples, we provide IoU and explanation F1 scores calculated using various thresholds for the model explanations (given their continuous nature) for a more comprehensive study. We vary the threshold of attention values in the model explanations and recalculate both IoU and F1 scores. A low threshold (e.g., 0.1) encompasses regions with minimal influence on the model's prediction results when calculating the degree of overlap with human annotations. Conversely, a high threshold (e.g., 0.9) focuses solely on regions with significant influence on the model's prediction results when determining the degree of overlap with human annotations. Using Figure~\ref{fig:heatmap} to better illustrate, the yellow-green areas in the model explanation may have an attention value of 0.1, while the red areas in the model explanation may have an attention value of 0.9. Figure~\ref{fig:All_Exp} displays the IoU and F1 scores achieved by each model across varying attention value thresholds from 0 to 0.9 (where attention values range from 0 to 1) for both Pancreas and LIDC datasets. In general, DUE has the highest IoU and F1 at different thresholds, and consistently outperforms others, evident in the red line consistently occupying a higher position in most cases. This superiority becomes more pronounced as the threshold increases, suggesting that DUE exhibits greater confidence in identifying important areas. This observation aligns with our findings in Section~\ref{subsec:QualitativeExp}.

\subsection{Sensitivity Analysis of Hyper-Parameter}

We evaluate the robustness of our proposed DUE framework to various changes in hyper-parameter $\lambda$ as shown in Equation~\ref {eq:due_objective}, which determines the balance between the predictive loss and explanation loss. Figure ~\ref{fig:sensitivity} shows that the PR-AUC is relatively stable across the range of $\lambda$ values, with a slight increase as $\lambda$ approaches 1. The IoU metric shows an initial increase with small values from 0.001 to 0.01 and the F1 score increases notably as $\lambda$ moves from 0.001 to 0.1, and then gradually declines, indicating a moderate emphasis on explanation loss leads to a better performance. Generally, our model outperformed the baseline model by a significant margin in both prediction accuracy as well as explainability. The optimal range for $\lambda$ is between 0.1 and 1, with the peak at 1, which suggests the overall performance is the best when the prediction loss and explanation loss are balanced.

\section{Conclusion} \label{sec:conclusion}
This paper introduces the Dynamic Uncertainty-aware Explanation supervision (DUE) framework, addressing challenges in applying explanation supervision to 3D medical images. Our approach overcomes issues such as altered spatial correlations, sparse 3D annotations, and varying uncertainty by introducing a diffusion-based 3D interpolation method with uncertainty-aware guidance. Through extensive experiments on diverse medical imaging datasets, we show that the DUE framework significantly improves the predictability and explainability of deep learning models in medical diagnosis, showcasing its potential to advance Explainable AI (XAI) in healthcare diagnostics.


\bibliographystyle{ACM-Reference-Format}
\bibliography{reference}

\appendix









\end{document}